\documentclass[11pt,a4paper]{article}
\usepackage[hyperref]{acl2019}
\usepackage{times}
\usepackage{multicol}
\usepackage{graphicx}
\usepackage{subcaption}
\usepackage{latexsym}
\usepackage{comment}

\usepackage{url}

\aclfinalcopy 

\usepackage{amsfonts}  
\usepackage{amsmath}  
\usepackage[linewidth=1pt]{mdframed} 
\usepackage{multirow}
\usepackage{graphicx}  
\usepackage[ruled]{algorithm2e}  

\newcommand\modelname{SummAE}

\newcommand\codeurl{\url{https://github.com/google-research/google-research/tree/master/summae}}     

\title{SummAE: Zero-Shot Abstractive Text Summarization using Length-Agnostic Auto-Encoders}

\author{
  Peter J. Liu\hspace{0.1cm}$^*$\\
  Google Brain \\
  \texttt{peterjliu@google.com}\\\And
  Yu-An Chung\hspace{0.1cm}\thanks{\hspace{0.2cm}Equal contribution}
  \hspace{0.1cm}\thanks{\hspace{0.2cm}Work done while interning
at Google Brain.}\\
  MIT CSAIL\\
  \texttt{andyyuan@mit.edu}\\\And
  Jie Ren\\
  Google Brain\\
  \texttt{jjren@google.com}\\
}

\date{}

\begin{document}
\maketitle


\begin{abstract}
    We propose an end-to-end neural model for zero-shot abstractive text summarization of paragraphs, and introduce a benchmark task, \emph{ROCSumm}, based on ROCStories, a subset for which we collected human summaries.
    In this task, five-sentence stories (paragraphs) are summarized with one sentence, using human summaries only for evaluation.
    We show results for extractive and human baselines to demonstrate a large \emph{abstractive gap} in performance. 
    Our model, \emph{\modelname}, consists of a denoising auto-encoder that embeds sentences and paragraphs in a common space, from which either can be decoded.
    Summaries for paragraphs are generated by decoding a sentence from the paragraph representations.
    We find that traditional sequence-to-sequence auto-encoders fail to produce good summaries and describe how specific architectural choices and pre-training techniques can significantly improve performance, outperforming extractive baselines.
    The data, training, evaluation code, and best model weights are open-sourced.
\end{abstract}

\section{Introduction}

\begin{figure}[!t]
\small
\begin{mdframed}
\textbf{(a) Story (paragraph) to summarize}\\
- Jason listened to the weather and heard it was going to be sunny. He thought the kids might like to go swimming. He gathered up the swimsuits, towels and sunscreen. Jason and the kids got into the truck and drove to the beach. They spent the next 2 hours playing and splashing in the surf.  \\
\\
\textbf{(b) Three human summaries}\\
- Jason saw a nice weather forecast and went to the beach with his kids for 2 hours. \\
- Jason took the kids swimming at the beach on a sunny day. \\
- Jason decided to take the kids to the beach since it was a sunny day.\\
\\
\textbf{(c) Best extractive sentence}\\
- Jason listened to the weather and heard it was going to be sunny. \\
\\
\textbf{(d) Unsupervised abstractive summary (ours)}\\
- Jason listened to the weather to be the coolest and fun day and he went to the beach and got ready.
\end{mdframed}

\normalsize

\caption{
(a) An example story to summarize from the validation set
for which we collected human summaries.
(b) Human summaries from independent raters.
(c) The extractive sentence with the highest ROUGE-1.
(d) Summary generated from our model without seeing any summaries.
}
\label{fig:nice_summary}

\end{figure}  


Extractive summarization has been studied extensively over the past several decades~\citep{gupta2010survey,ferreira2013assessing}.
However, humans typically summarize abstractively, paraphrasing and performing non-trivial compression of details that are difficult to encode in classical summarization algorithms.
Recent progress in neural language models~\citep{sutskever2014sequence,jozefowicz2016exploring,radford2019language} has enabled models to generate near fluent language that are not mere regurgitations of training data.
With large datasets of (document, summary) pairs primarily from the news domain, abstractive summarization has been approached as a supervised neural sequence transduction problem~\citep{rush2015neural,nallapati2016abstractive,narayan2018don,fabbri-etal-2019-multi}.
Outside of news, however, such large parallel datasets are rare, due to the cost-prohibitive ``labeling'' process (i.e., reading long documents and writing summaries).
Much more commonly available are large corpora of documents without summaries.
It is therefore desirable to have models capable of automatically summarizing documents abstractively with little to no supervision.

In contrast to abstractive methods, many extractive approaches do not rely on example summaries. Inspired by that we study the extreme case of no exposure
to summaries during training, or unsupervised (zero-shot) abstractive summarization (UAS).

Recently there has been some, thus far limited, work on UAS for both the multi-document~\citep{chu2019meansum} and single-document~\citep{isonuma2019unsupervised} cases.
In this work we focus on UAS of paragraphs with a sentence, which is perhaps the most basic form of multi-sentence single-document summarization.
In particular, we summarize the five-sentence stories from ROCStories~\citep{mostafazadeh2016corpus} and show that there is a non-trivial \textit{abstractive gap} between human and extractive performance, making it a suitable benchmark for measuring progress in UAS.

Our approach is based on training a denoising auto-encoder~\citep{vincent2008extracting} that encodes sentences and paragraphs in a shared space.
The decoder input is pre-pended with a special token to signal whether to decode a sentence or a paragraph, and a summarizing sentence is generated by decoding a sentence from an encoded paragraph.
However, we found that traditional approaches to training such an auto-encoder resulted in non-overlapping sentence and paragraph latent sub-spaces --- which we call \emph{segregation} --- resulting in long, multi-sentence summaries.
We describe architectural modifications and self-supervised pre-training objectives to prevent segregation and improve performance significantly beyond sentence-extractive baselines.
While the goal of human performance is still far, we believe the techniques presented here are a major step in that direction.

In summary, our contributions are as follows.
\begin{enumerate}
  \item  We introduce a new benchmark task, \emph{ROCSumm}, for measuring progress toward human performance on UAS.
  \item We propose a novel end-to-end, fully-differentiable neural model for UAS of paragraphs.
  \item We describe novel self-supervised (pre-training) and denoising objectives that significantly improve performance beyond sentence-extractive baselines.
  \item We conduct ablation experiments showing the importance of architectural choices and model objectives.
\end{enumerate}


\section{A new task for unsupervised abstractive summarization}


Our new task re-purposes and augments an existing dataset, ROCStories~\citep{mostafazadeh2016corpus,mostafazadeh2017lsdsem}, originally designed for the ``Story Cloze Test'' (SCT), where a model must choose the correct fifth sentence of two candidates given the first four.
The stories are
self-contained, diverse, realistic, non-technical, high-quality, and have a coherent story arch.
The human performance on the SCT task is close to 100\%.

Our proposed UAS task involves summarizing the five-sentence training ROCStories
with a single sentence without summaries at training, i.e., perform zero-shot
summarization.
We found summaries by independent human raters have high similarity in this task, suggesting it is well-defined, and relatively unambiguous, in contrast to other summarization tasks where the desired length or the topic of the summary is unclear.
The simplicity of the task is conducive to iterating quickly and making rapid progress in UAS.
Having only five-sentences and a low-bound on total number of words avoids engineering issues that often arise with very long sequences.
Due to the constraints, it is simple to calculate the maximum (sentence) extractive summarization performance, which is far from the human performance (see Table~\ref{tab:main_results}) suggesting a need for abstractive models.
In contrast, it is unclear for example, what human performance is on the popular CNN/DailyMail (supervised) summarization task~\citep{see2017get} and whether abstractive models provide much of a benefit over extractive ones on it~\citep{kryscinski2019neural}.

\subsection{Collection of reference summaries for evaluation}
To evaluate summarization models, we collected multiple summaries from independent, experienced, and highly-reputable Amazon Mechanical Turk (AMT) workers.
The full worker selection criteria and AMT template can be found in Appendix~\ref{app:amt}.
We split the original 98,159 training stories into 95\% train, 5\% valid, and 5\% test examples, and collect 3 human summaries each for 500 validation examples and 500 test examples.
Collecting multiple summaries allowed us to estimate human performance as well as treat multiple ``right'' answers more fairly by averaging metrics across the summaries for a given example.
An example story with 3 human summaries, the best extractive sentence, and one of our model summaries can be found in Figure~\ref{fig:nice_summary}.


\section{Related Work}
\label{sec:related}
\citet{chu2019meansum} proposed a model for zero-shot multi-document abstractive summarization, where the mean of the representations from an auto-encoder for input documents is used to decode a summary.
\citet{isonuma2019unsupervised} proposed to summarize a product review by describing it as a discourse tree, where the summary is the root and the child sentences explain their parent.

\citet{baziotis2019seq} performed sentence compression by chaining two sequence-to-sequence (seq2seq) models as an auto-encoder.
A Straight-Through Gumbel-Softmax estimator~\citep{jang2017categorical} was used to sample an output sequence from the first seq2seq, which was encouraged to be language via a pre-trained language model loss.
It was also encouraged to be related to the original sentence by using it as input to the second seq2seq model, which was trained to reconstruct the original sentence.
\citet{fevry2018unsupervised} used a denoising auto-encoder for sentence compression as well.
An input sentence was artificially extended and word-shuffled, encouraging the model to learn to exclude and compress, producing shorter sentences.

\citet{wang2018learning} trained a Cycle-GAN model \citep{zhu2017unpaired} to learn a document-to-summary mapping given large datasets of unpaired documents and summaries. However,
due to the model being exposed to summaries during training, it is not zero-shot summarization.
Further, unlike the original Cycle-GAN model on images, it is non-differentiable since the discriminator must distinguish real from generated in the discrete language domain, and relies on REINFORCE~\citep{williams1992simple}.

\citet{radford2019language} trained a large language model on a large Web text dataset and found that
the model could produce zero-shot summaries if prompted with a document followed by \textit{TL;DR}, though they considered them rudimentary and unusable.

Historically there have been strong parallels in the development of neural sequence transduction models for translation and summarization, relying
on some flavor of sequence-to-sequence learning. We  depart significantly from recent unsupervised translation work \citep{lample2018unsupervised,lample2018phrase,artetxe2018unsupervised} where models are exposed to both source and target sequences, though unpaired.  In our work,
models must learn to produce target sequences (i.e., summarize) having only been exposed to source sequences (documents)
during training. 





\newcommand\enc{\phi_{enc}}
\newcommand\dec{\phi_{dec}}
\newcommand\paragraphs{\mathbb{P}}
\newcommand\encparam{\theta_{E}}
\newcommand\decparam{\theta_{G}}
\newcommand\texts{\mathbb{T}}
\newcommand\Reals{\mathbb{R}}
\newcommand\Zspace{\Reals^{z_{dim}}}
\newcommand\sind{\beta}

\section{Model and Methods}
\label{sec:model}

\subsection{Architecture}
\label{sec:architecture}
\begin{figure}[t!]
  \begin{center}
    \includegraphics[width=1.0\columnwidth]{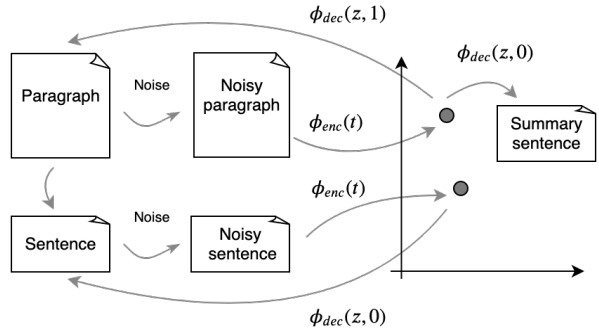}
    \caption{Model architecture of \modelname, whose backbone is a denoising auto-encoder. The encoder $\phi_{enc}$ maps paragraphs and sentences into a common space, from which either can be decoded via the decoder $\phi_{dec}$ by conditioning on two different beginning-of-sequence tokens $\beta\in \{0, 1\}$. During training, we add noise to text sequences before mapping to encourage $\phi_{enc}$ to learn more robust text representations.}
    \label{fig:model}
  \end{center}
\end{figure}

Our summarization model, \modelname, is depicted in Figure~\ref{fig:model} and consists of a denoising auto-encoder, $(\enc, \dec)$, capable of auto-encoding text sequences $\texts$ that can be sentences or paragraphs.
In particular, the encoder, $\enc: \texts \mapsto \Zspace$, is a deterministic function, parameterized by $\encparam$, mapping text to a latent vector representation, $\enc(t)=z \in \Zspace$.
For an input sequence $t\in\texts$, we add random noise to $t$ (described in Section~\ref{sec:noise}), $\tilde{t} = \eta(t)\in\texts$.
We consider two encoder implementations:
\begin{enumerate}
  \item a bidirectional RNN~\citep{schuster1997bidirectional} where $\enc(t)=z$ is derived from the RNN's final hidden state, $h = [h_{\rightarrow}, h_{\leftarrow}] \in \Reals^{2h_{dim}}$, followed by an affine transformation:
  \begin{align*}
    h &= \mathrm{RNN}_{enc}(\tilde{t}),\\
    z &= Wh + b;
  \end{align*}
  \item a Transformer stacked with $N$ identical Transformer encoder blocks (denoted as $\mathrm{TRF}_{enc}$) from~\citet{vaswani2017attention}. $\enc(t)=z$ is derived from the output representation of the first token as in~\citet{devlin2019bert}, followed by an affine transformation:
  \begin{align*}
    h_0 &= \tilde{t},\\
    h_l &= \mathrm{TRF}_{enc}(h_{l-1}), \forall l\in [1, N],\\
    h &= h_N[0, :],\\
    z &= Wh + b.
  \end{align*}
\end{enumerate}

The decoder, parameterized by $\decparam$, is an auto-regressive generative model defining a probability distribution over $\texts$ conditioned on $z$. 
We also condition the decoder on whether to decode a sentence or paragraph, using two different beginning-of-sequence tokens indicated by $\sind\in\{0, 1\}$.
The reconstructed input, $\hat{t}$, is obtained by sampling one token at a time until a special end-of-sequence token is obtained:
\begin{align}
  \hat{t} &= \dec(z, \sind) \nonumber\\
  &\sim p_{\decparam} (\hat{t} | z, \sind) \nonumber\\
  &= \prod_i p_{\decparam} (\hat{t}_i | \hat{t}_1,..., \hat{t}_{i-1}, z, \sind).
  \label{eq:decoding}
\end{align}
We consider two decoder implementations:
\begin{enumerate}
  \item a unidirectional RNN that conditions on $z$ by concatenating the decoder input embeddings with $z$ at each time-step;
  \item a Transformer with causal, masked attention that conditions by adding $z$ to each input embedding. This is similar to the Transformer decoder in~\citet{vaswani2017attention} without decoder-encoder attention.
\end{enumerate}
In both cases, we avoid decoder-encoder attention to encourage all semantic information to be encapsulated in $z$. 

In our dataset, a single example is a paragraph consisting of $n(p)$ sentences, $p=(s_1, ..., s_{n(p)})$, and the auto-encoder contributes two reconstruction loss terms, one for the sentences and one for the paragraph, weighted by $\lambda_s$ and $\lambda_p$.
\begin{multline}
  \ell_{rec}(p, \encparam, \decparam)\\
  = \lambda_s \frac{1}{n(p)}\sum_{i=1}^{n(p)}\ell_{ce}(s_i, \hat{s_i}; \encparam, \decparam)\\
  + \lambda_p \ell_{ce}(p, \hat{p}; \encparam, \decparam),
  \label{eq:rec}
\end{multline}
where $\ell_{ce}(x, \hat{x})$ is the standard cross-entropy loss between the input sequence and its reconstruction.
We optimize it using gradient descent and teacher-forcing~\citep{williams1992simple}.

Our approach to summarize paragraphs with a sentence is to prompt the decoder to generate a sentence (using $\sind=0$) conditioned on the latent vector $z$ of a paragraph.
However, simply training an RNN or Transformer auto-encoder as described generally fails (as we see in Section~\ref{sec:results}); we hypothesize that the encoder can learn to map sentences and paragraphs to separate regions in the latent space, and the decoder can recognize whether to decode a sentence or paragraph in reconstruction based solely on the location of $z$ and ignore $\beta$.
We find this can result in the decoder generating a paragraph even if prompted for a sentence.
We call this phenomenon \emph{segregation}.



Ideally, the auto-encoder learns a higher-level latent \emph{concept} conveyed in paragraphs and sentences, \textit{disentangled} from their original expression as paragraphs and sentences.
To explicitly encourage this we investigated adding an adversarial discriminator/critic $D$, which is trained to classify whether a latent vector $z$ is an encoded sentence or a paragraph.
In other words, it learns $p_D(y(t) | \enc(t))$, where $y(t)$ is 0 if $t$ is a sentence and 1 if a paragraph, while the auto-encoding loss is augmented to encourage fooling the discriminator into classifying paragraphs as sentences.
Similar approaches have been used in style transfer~\citep{hu2017toward,shen2017style,romanov2019adversarial} and unsupervised machine translation~\citep{lample2018unsupervised}, although not for abstractive summarization.
Details of our implementation can be found in Appendix~\ref{app:critic}.

In our experiments we found adding the critic was very effective for generating one-sentence short summaries.
However, for some encoder-decoder configurations, the critic was found to be unnecessary and even harmed performance, which we discuss in Section~\ref{sec:results}.

\subsection{Adding noise to text}
\label{sec:noise}
We use a denoising auto-encoder rather than a standard auto-encoder by reconstructing a text sequence from a noisy version of it.
Denoising can be seen as a useful self-supervised objective for improving representations as seen in~\citet{devlin2019bert}; it also serves as a form of data augmentation, effectively increasing the number of training examples; finally, it discourages merely learning the identity function without having to reduce the information bottleneck $z_{dim}$ to a very small value.
We employ two techniques for adding noise:

\newcommand\MASK{\texttt{MASK}}

\paragraph{Randomly masking tokens}
Similar to~\citet{devlin2019bert} and~\citet{song2019mass}, we randomly mask the input sequence tokens at training before feeding it to the encoder.
However, instead of only predicting masked tokens, we generate the full denoised sequence.
We apply the masking with the following procedure:
\begin{enumerate}
  \item Select sequences to mask with probability $p_s < 1.0$, so that some of the sequences are unpermuted as they are during test time.
  \item For selected sequences, replace each token with a \textless \MASK\textgreater\ token with probability $p_m$.
\end{enumerate}

\paragraph{Permuting order of sentences within paragraphs}
Even with token-masking we observed a failure mode where the latent representation of a paragraph overly focuses on the first sentence of the paragraph and memorizes it.
Indeed, it is the best sentence to memorize for the purpose of reconstructing a paragraph.
However, to encourage learning the structure and coherence of paragraphs beyond the first sentence, with probability $p_{perm}$, we permute the order of sentences in a paragraph and train the auto-encoder to recover the original paragraph.


\subsection{Pre-training of encoder and decoder}
\label{sec:pretraining}

\newcommand\SEP{\texttt{SEP}}
\newcommand\A{\texttt{A}}
\newcommand\B{\texttt{B}}

Motivated by the recent success in self-supervised language representation learning~\citep{peters2018deep,howard2018universal,radford2018improving,devlin2019bert,song2019mass,zhang2019ernie,yang2019xlnet}, we propose several strategies that pre-train the encoder and decoder before optimizing them jointly with the auto-encoding objective~(Equation~\ref{eq:rec}).
The strategies are applied jointly in the pre-training phase by adding the corresponding losses.

Although we adopt the paradigm of pre-training followed by fine-tuning, there are two significant differences with other work.
In past work labeled data are available for the downstream tasks and fine-tuning is supervised; in our work, however, both pre-training and fine-tuning are unsupervised.
Additionally, most previous work pre-trained their models on extremely large corpora different from their downstream datasets, whereas our model learns everything from the same dataset%
\footnote{We tried pre-training on large corpora, but it didn't help.}.

\paragraph{Encoder pre-training \#1: Corrupted Paragraph Prediction}
Summarizing a paragraph requires understanding how its sentences follow each other to form a coherent narrative.
To encourage good paragraph representations we propose a novel pre-training task classifying whether a paragraph has been \emph{corrupted} by swapping the order of the $i$-th sentence with its next sentence.
We implemented this by adding a logistic regression layer to the encoder paragraph output $z$, and minimizing cross-entropy error with 50\% of paragraphs corrupted, and 50\% unmodified.
We refer to this task as Corrupted Paragraph Prediction (CPP).

\paragraph{Encoder pre-training \#2: Next Sentence Prediction}
We propose another pre-training objective encouraging the encoder to understand how sentences follow within a paragraph.
The objective, referred to as Next Sentence or Same Paragraph (NSSP), shares an idea with BERT's Next Sentence Prediction (NSP) objective ~\citep{devlin2019bert}---to classify whether two sentences are adjacent---but is modified to be more difficult. 
As in BERT, we sample the sentence pairs \A\ and \B\ such that 50\% of the time \B\ follows \A, otherwise it does not (negatives). However, negative pairs are sampled from within the same paragraph instead of from the whole corpus. This is more challenging as sentences from the same paragraph usually are more similar and about the same topic. We observed this harder negative sampling leads to better downstream summarization performance.

Further our implementation differs in the following way from BERT's NSP:
\begin{itemize}
  \item In BERT, each input sequence is constructed by concatenating \A\ and \B\, separated by a \textless \SEP\textgreater\ token. To better differentiate \A\ and \B, BERT further employs segment embeddings. In contrast, we encode \A\ and \B\ independently, avoiding the need for the separator token and segment embeddings.
  \item In BERT, a binary multi-layer perceptron is added during pre-training. Instead of introducing these extra parameters we directly take the dot product of the two sentence representations followed by a sigmoid function: 
  \begin{equation*}
    \sigma(\enc(\A) \cdot \enc(\B)), \A, \B \in \texts.
  \end{equation*}
  This also restricts the capacity of the classifier, forcing all relevant features into $z$. This implementation is simpler, and more generic since it is not tied to a specific encoder (e.g., BERT/Transformer) architecture.
\end{itemize}
In contrast to CPP, which learns to encode the sentence relationships in paragraph representations, NSSP is taught to encode sentence relationships in individual sentence representations.

\paragraph{Decoder pre-training: Auto-regressive Language Modeling}
We pre-train the decoder with a standard auto-regressive language modeling (LM) objective similar to \citet{ramachandran2017unsupervised}.
We implement this by setting $z=0$ in Equation~\ref{eq:decoding} regardless of the input sequences in pre-training, so that the decoder receives no conditioning signal during teacher-forcing, making the process equivalent to a standard LM task.

\begin{figure*}
  \centering
  \begin{subfigure}[t]{.6\columnwidth}
    \includegraphics[width=\linewidth]{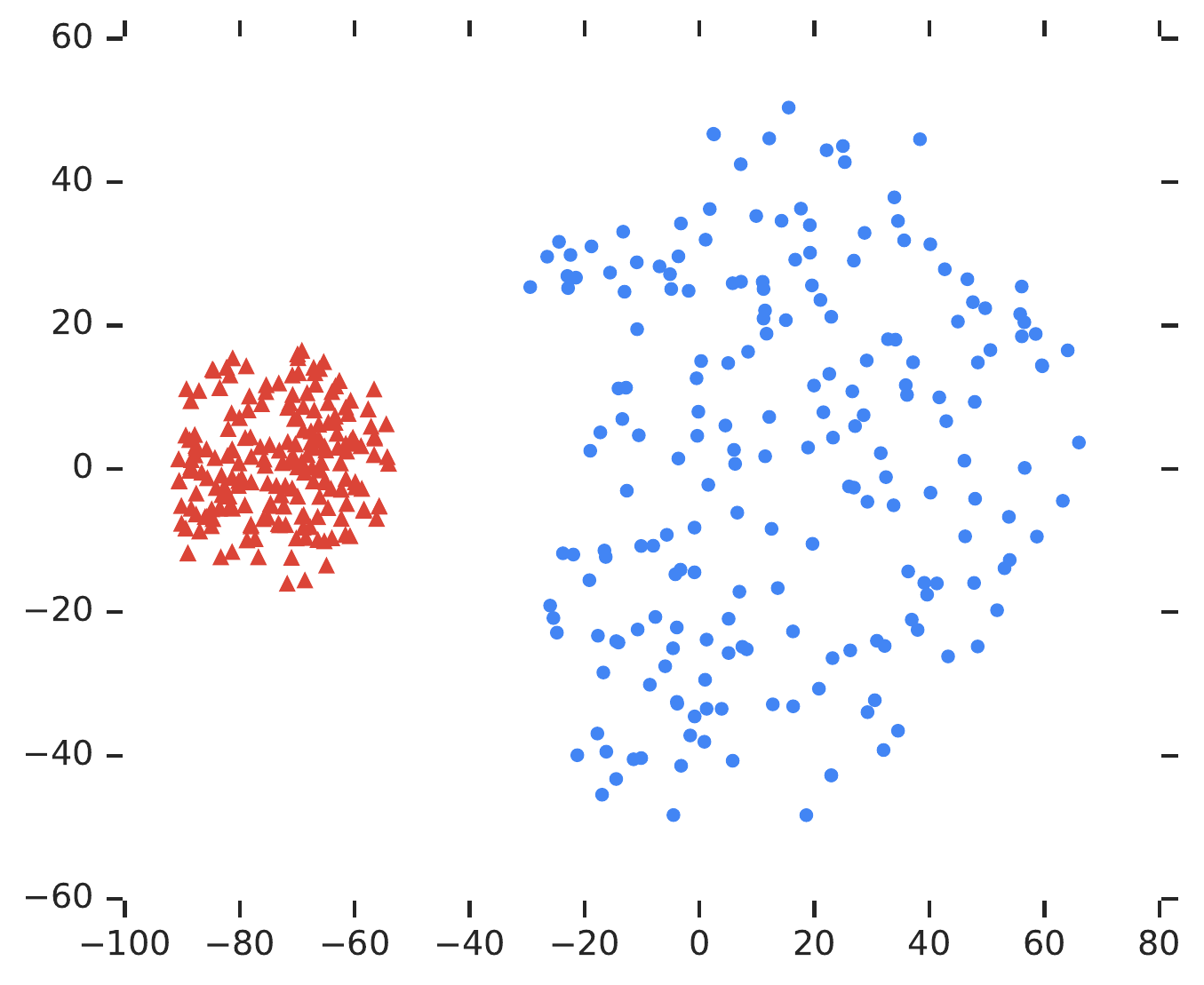}
    \caption{RNN-RNN without critic.}
    \label{fig:latent_rnn_nocritic}
  \end{subfigure}
  \begin{subfigure}[t]{.6\columnwidth}
    \includegraphics[width=\linewidth]{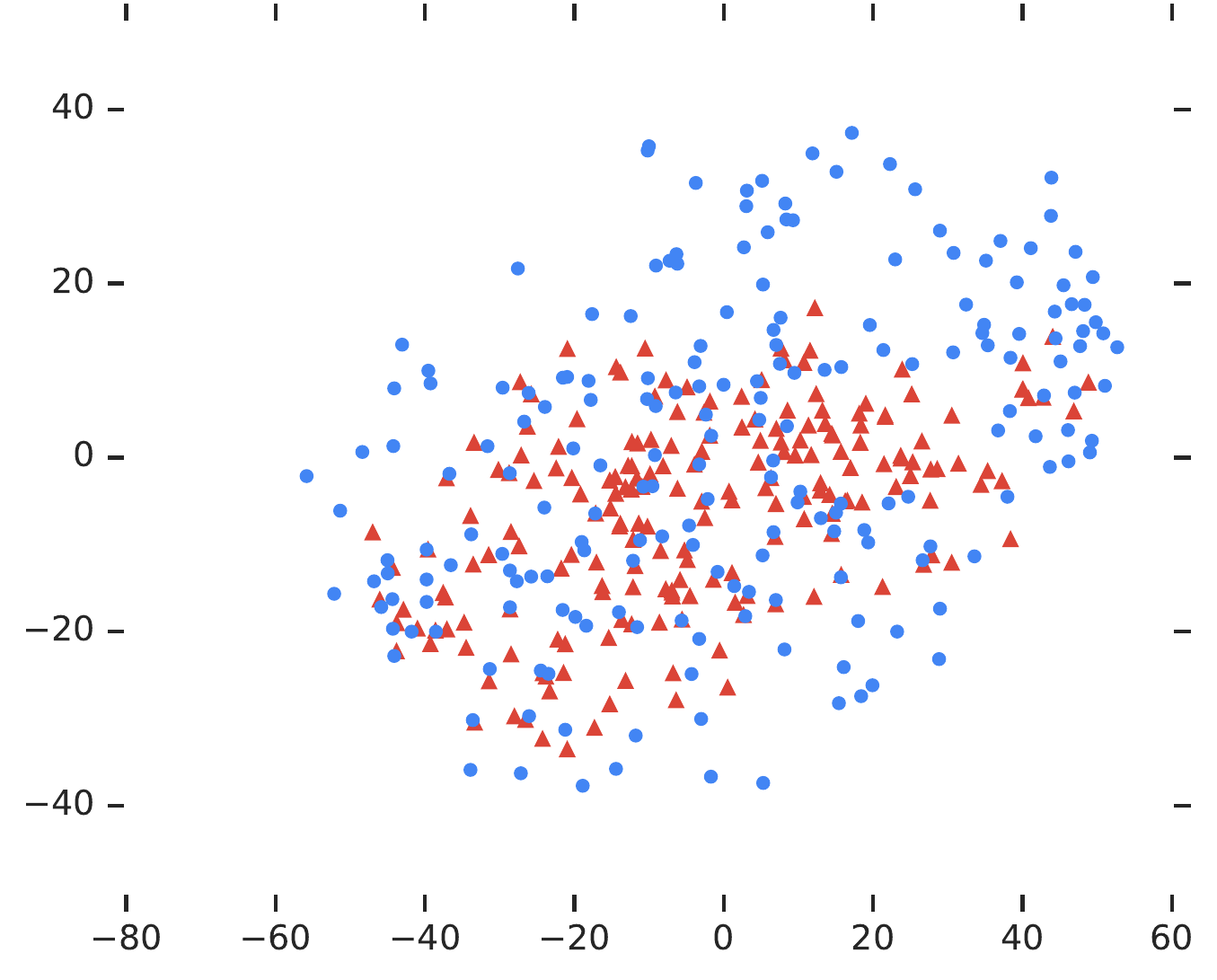}
    \caption{RNN-RNN with critic.}
    \label{fig:latent_rnn_critic}
  \end{subfigure}
  \begin{subfigure}[t]{.6\columnwidth}
    \includegraphics[width=\linewidth]{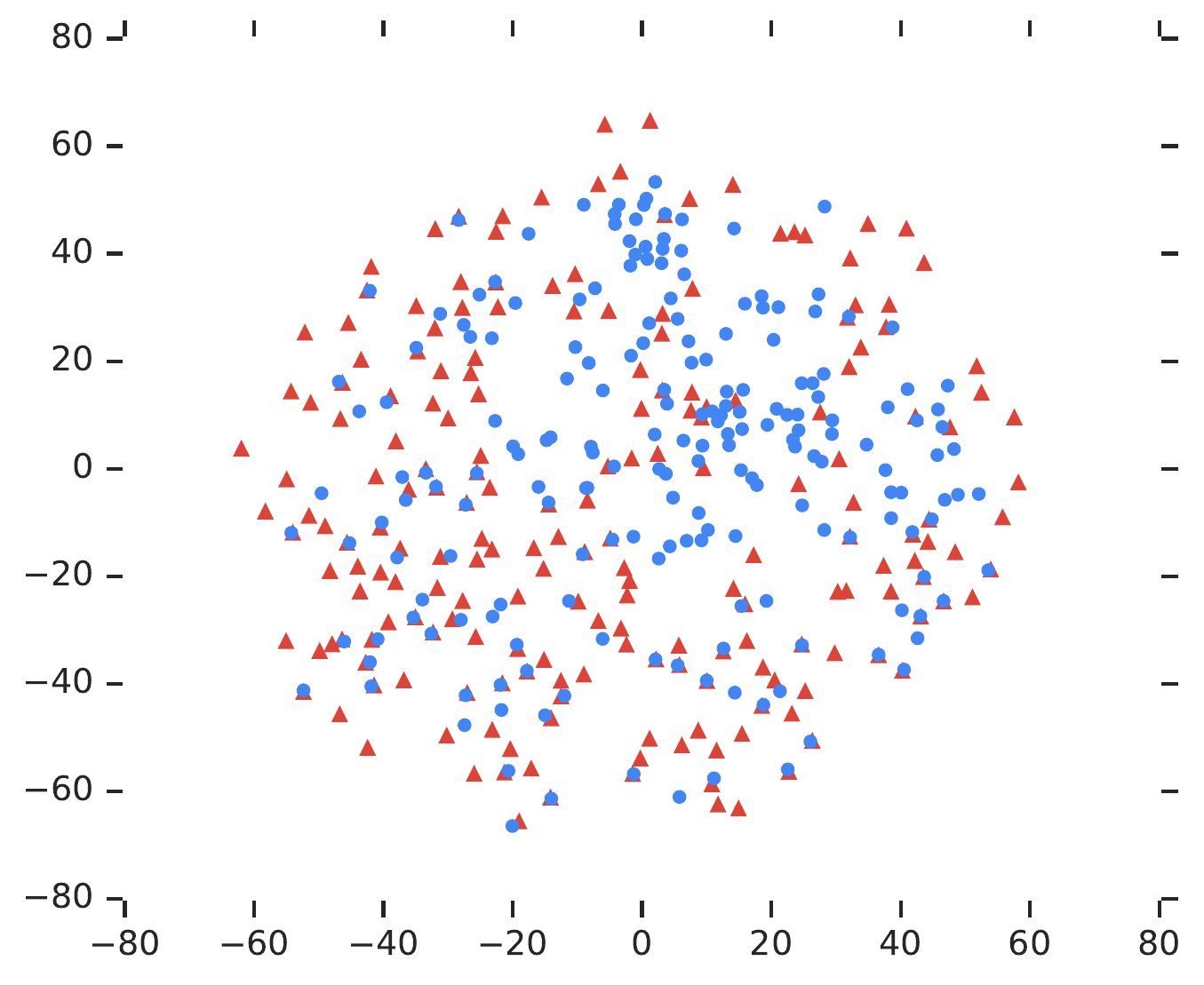}
    \caption{TRF-RNN with LM pre-training and without critic.}
    \label{fig:latent_trf_lm}
  \end{subfigure}
  \caption{2-D t-SNE visualizations for \modelname\ latent space. Each blue circle corresponds to a sentence and each red triangle corresponds to a paragraph (best viewed in color).}
\end{figure*}

\section{Experiments}
\subsection{Metrics}
We found human summaries, which are quite abstractive, to be unfairly punished on ROUGE-2 and ROUGE-X precision variants; and thus only report ROUGE-1 and -L recall scores~\citep{lin2004rouge}.
We truncated generated summaries to their first sentence and to a word limit of 20 to prevent preferring models with overly long summaries, similar to~\citet{rush2015neural}, although they used a byte limit.
The limitations of ROUGE are well-known, and in particular favor extractive methods, since paraphrasing and synonyms are not rewarded fairly.
However, at the time of writing, ROUGE remains a dominant metric in summarization research~\citep{kryscinski2019neural}.
We treated the 3 summaries per evaluation example equally and report average ROUGE scores over the 
500 test examples, using the 500 validation examples for hyper-parameter tuning. We observed little difference in averaged validation and test scores.

\subsection{Baselines}

\paragraph{Extractive}
Because the documents consist of only five sentences, for each example there are only five possible sentence-extractive summaries.
Thus we computed the performance of selecting sentence $i$ as the summary, for $i\in \{1, ..., 5\}$, and denote these as \emph{Extract\_$i$}.
We also computed maximum achievable sentence-extractive score, by selecting the best performing sentence per example evaluated based on ROUGE-1 (recall) against the 3 human summaries, which we refer to as \emph{Extract Oracle}.
Although this method cheats by looking at the test data and may not actually be achievable in practice, it is useful to have an estimate of the ceiling of extractive methods.

\paragraph{Human}
We estimated human performance by computing (a) the maximum and (b) the average ROUGE between all pairs of human summaries
for each evaluation example, and took the mean across examples.

\paragraph{MeanSum-single}
Although designed for multi-document summarization, we adapted MeanSum~\citep{chu2019meansum} for the single-document case by treating each 
sentence as a document similar to \citet{isonuma2019unsupervised}, where it is called MeanSum-single. Since MeanSum generates summaries of the same shape as the input documents, this adaptation generates sentence-summaries.

\subsection{Training details and hyper-parameters}
\paragraph{Text Tokenization}
We used a data-driven sub-word tokenizer~\citep{sennrich2016neural} with a vocabulary size of $2^{15}$ to convert text into integer ID tokens,
followed by a 128-dim embedding layer.
We shared the input and output embedding layers as in~\citet{press2017using}.

\paragraph{Architecture}
We experimented with three base encoder-decoder configurations for \modelname: RNN-RNN, TRF-TRF, and TRF-RNN, where TRF stands for Transformer.
Both $\mathrm{RNN}_{enc}$ and $\mathrm{RNN}_{dec}$ were single-layer GRUs~\citep{chung2014empirical} with $h_{dim}=512$, while $\mathrm{RNN}_{enc}$ was bidirectional and $\mathrm{RNN}_{dec}$ was unidirectional.
$\mathrm{TRF}_{enc}$ was a stack of 2 Transformer encoder layers with each consisting of a self-attention sub-layer with 8 attention heads and a point-wise feed-forward network with 512 hidden units.
$\mathrm{TRF}_{dec}$ had the same hyper-parameter setting as $\mathrm{TRF}_{enc}$ but was composed of Transformer decoder layers.
$z_{dim}$ was set to 256.
All model weights, including embeddings were initialized randomly. Decoding at each time-step was performed greedily using arg-max.


\begin{table*}[t]
	\caption{ROUGE recall scores and generated summary lengths (number of words and sentences) for estimated human performance, extractive baselines, MeanSum-single, and variants of \modelname. By default, all \modelname\ models incorporate token masking and paragraph shuffling without critic and any pre-training. TRF stands for Transformer. Numbers in bold denote the best performed models in each category based on ROUGE; numbers in italics denote the models that are not qualified for generating one sentence summaries. Superscript letters, \textsuperscript{a,b,c,d}, denote model pairs that are compared in human evaluations in Table~\ref{tab:human}.}
\label{tab:main_results}
\small
\begin{center}
\begin{tabular}{ll|ccccc}

\multicolumn{1}{c}{}
&\multicolumn{1}{l|}{\bf Model}
&\multicolumn{1}{c}{\bf ROUGE-1}
&\multicolumn{1}{c}{\bf ROUGE-L}
&\multicolumn{1}{c}{\bf Num words}
&\multicolumn{1}{c}{\bf Num sentences}
\\

\hline
\hline
\rule{0pt}{2.5ex}  &  Human average  &  45.0  &  37.7 & 13.9 & 1.0\\
&  Human maximum\textsuperscript{d}  &  \textbf{52.7}  &  \textbf{44.1} & 17.6 & 1.0\\

\hline
	\multirow{6}{*}{\rotatebox[origin=c]{90}{\emph{Extractive}}}
\rule{0pt}{2ex}	&  Extract\_1  &  27.3  &  24.8  &  7.9  &  1.0\\
	&  Extract\_2  &  19.5  &  16.7  &  8.9  &  1.0\\
	&  Extract\_3  &  19.4  &  16.4  &  9.0  &  1.0\\
	&  Extract\_4  &  20.5  &  17.7  &  8.7  &  1.0\\
	&  Extract\_5  &  24.6  &  20.9  &  9.3  &  1.0\\
	&  Extract Oracle  &  \textbf{36.7}  &  \textbf{31.9}  &  9.7  &  1.0\\

\hline
	\multirow{15}{*}{\rotatebox[origin=c]{90}{\emph{Abstractive}}}
	\rule{0pt}{2.5ex}&  \modelname\ RNN-RNN  &  \textit{33.7}  & \textit{27.2}  &  \textit{41.2}  &  \textit{4.2}\\

	&  + critic\textsuperscript{a,b}    &  19.8  &  17.4  &  9.5  &  1.0\\
	&  \quad + LM pre-training\textsuperscript{b,c}  &  26.4  &  21.6  &  11.5  &  1.0\\

\cline{2-6}

  \rule{0pt}{2.5ex}  &  \modelname\ TRF-TRF  &  \textit{22.4}  &  \textit{20.3}  &  \textit{44.1}  &  \textit{5.1}\\
    &  + critic  &  22.9  &  19.9  &  9.4  &  1.0\\
    &  \quad + LM pre-training &  25.8  &  21.2  &  10.2  & 1.0\\
    
\cline{2-6}
  \rule{0pt}{2.5ex}  &  \modelname\ TRF-RNN &  \textit{26.5}  &  \textit{22.8}  &  \textit{39.7}  &  \textit{3.0}\\
    &  + critic     &  23.2  &  19.9  &  10.4  &  1.0\\
    &  \quad + LM pre-training  &  26.4  &  22.7  &  10.7  & 1.0\\
    &  + LM pre-training  &  33.4 &  27.5  &  16.3  &  1.0\\
    &  \quad - token masking  &  30.9  &  25.1  &  16.5  &  1.0\\
    &  \quad - paragraph shuffling &  31.8  &  26.5  &  14.7  &  1.0\\
    &  \quad + CPP pre-training  &  34.4  &  28.0  &  19.6  & 1.0\\
    &  \quad + NSSP pre-training\textsuperscript{c,d}  &  \textbf{36.5}  &  \textbf{29.3}  &  19.4  &  1.0\\
    
\cline{2-6}
	\rule{0pt}{2.5ex}  &  MeanSum-single\textsuperscript{a}    & 15.6 & 13.4 & 9.5 & 1.0 \\
    &  \quad - token masking \quad &  12.0 & 10.5 & 9.4 & 1.0 \\
\end{tabular}
\end{center}
\end{table*}

\paragraph{Optimization} We performed gradient descent using the Adam optimizer~\citep{kingma2015adam} with a learning rate of 0.001 and batch size of 64.
The models were trained with early-stopping, using maximum validation ROUGE-1 (recall).
The number of pre-training steps was set to 100,000, when pre-training was employed.

When critic/discriminator was used, it was implemented as a multi-layer perceptron with 128 hidden units.
When token mask was added, we set $p_s=0.15$ and $p_m=0.8$.
For permuting sentences in paragraphs, we set $p_{perm}=0.5$.
The critic and noise were only added during the fine-tuning phase.

We experimented with different hyper-parameters for constructing \modelname\ but found the reported setting worked the best empirically.



\subsection{Results and discussion}
\label{sec:results}
Table~\ref{tab:main_results} shows ROUGE scores and summary lengths for human and extractive baselines, and for \modelname\ with enhancements described in Section~\ref{sec:model}.
For the base encoder-decoder configurations (RNN-RNN, TRF-TRF, and TRF-RNN), token masking and paragraph shuffling were added as noise, but no pre-training was done and no critic was added.

\paragraph{Human and extractive baseline performance}
The best extractive sentences were unsurprisingly the first and last, as the sentences introducing the subject, and revealing the ending.
Extractive Oracle is by definition the best and its performance was considerably higher than any fixed-index sentence to extract.
Estimated human performance was much higher than even the Extractive Oracle, suggesting an abstractive gap.

\paragraph{Critic effectively restricts summary-length}
All three base encoder-decoder configurations tended to generate overly long/invalid summaries as shown by the number of words (close to 40) and sentences (exceeding 1), indicating the $\beta$ in Equation~\ref{eq:decoding} was likely  ignored.
Once the critic was added, the model was able to generate single-sentence short summaries.

To validate our hypothesis that \emph{segregation} in the latent space was the underlying problem causing long summaries, we visualized the latent space of
\modelname\ RNN-RNN (similar plots for TRF-TRF and TRF-RNN) in 2-D using t-SNE~\citep{maaten2008visualizing}, with and without the critic.  In Figure~\ref{fig:latent_rnn_nocritic} where there was no critic, sentence and paragraph representations indeed were mapped to completely separate regions; while in Figure~\ref{fig:latent_rnn_critic} the adversarial feedback from the critic effectively merged the two clusters, supporting our hypothesis.

\paragraph{Effect of LM pre-training}
As can be seen from the table, LM pre-training always improved the model performance while keeping the summaries short: for all of three encoder-decoder configurations, it boosted their ROUGE-1 scores from 19.8 to 26.4, 22.9 to 25.8, and 23.2 to 26.4, respectively.
Qualitatively, we found models with LM pre-training generated more fluent summaries as well.

\paragraph{Best encoder-decoder configuration}
Interestingly, we observed that TRF-RNN outperformed the other two configurations that use the same sequential architecture for both encoder and decoder.
Similar results were reported in~\citet{chen2018best}, where they found their \emph{hybrid} model composed of a Transformer encoder and RNN decoder worked the best in machine translation. Another possible reason is that the decoder here does not need to attend to long-term dependencies, which is where Transformers have major advantages over RNNs.

Surprisingly we found that the TRF-RNN variant enhanced with LM pre-training did not require a critic to generate one-sentence summaries and prevented segregation as the other encoder-decoder configurations did.
Figure~\ref{fig:latent_trf_lm}, visualizes its corresponding latent space where the paragraph and sentence representations were mapped to the same region, suggesting that LM pre-training has the same effect of critic in this specific configuration.
It also outperformed the model with both critic and LM pre-training (33.4 vs. 26.4).

\paragraph{Effect of token masking and paragraph shuffling}
By default both token masking and paragraph shuffling introduced in Section~\ref{sec:noise} were added as noise to base \modelname\ models during auto-encoding.
Removing either one degraded the performance: ROUGE-1 score decreased from 33.4 to 30.9 when token masking was removed and to 31.8 when paragraph shuffling was removed.
During experimentation, we also found adding this noise produced more stable results across different runs.

\paragraph{Effect of NSSP \& CPP pre-training}
\modelname\ TRF-RNN with only decoder pre-training (33.4) already surpassed the best extractive sentence (Extract\_1: 27.3).
With encoder pre-training with either NSSP or CPP, we observed a further improvement from 33.4 to 36.5 and to 34.4, respectively.

Our best \modelname\ model---constructed by a Transformer encoder and an RNN decoder---was pre-trained with NSSP and LM objectives, followed by denoising auto-encoding of masked and shuffled input sequences.
It achieved (ROUGE-1, ROUGE-L) = (36.5, 29.3) that significantly outperformed the fixed-index extractive sentences and was comparable with Extract Oracle, which looks at the test data.







\paragraph{Human evaluation of model summaries}
\label{sec:human_eval}
\begin{table*}[ht]
\caption{Human evaluation results comparing model summary pairs side-by-side on fluency and information relevance. Superscript letters, \textsuperscript{a,b,c,d},
	correspond to models in Table~\ref{tab:main_results}. The * denotes statistical significance 
	at $p<0.001$ using a Binomial Two-Tailed test
	(null hypothesis: the models are equally good).}
\label{tab:human}
\small
\begin{center}
\begin{tabular}{l|c|c}

\multicolumn{1}{l|}{\bf Model preference}
&\multicolumn{1}{r|}{\bf Fluency}
&\multicolumn{1}{r}{\bf Information }  \\

\hline
\textsuperscript{a} RNN-RNN + critic $\geq$ MeanSum-single + noise & 77\%\textsuperscript{*} & 80\%\textsuperscript{*}   \\
\textsuperscript{b} RNN-RNN + critic + LM $\geq$   RNN-RNN + critic & 80\%\textsuperscript{*} & 80\%\textsuperscript{*}   \\
\textsuperscript{c} TRF-RNN + LM + NSSP $\geq$ RNN-RNN + critic + LM & 45\% & 86\%\textsuperscript{*} \\
\textsuperscript{d} Human $\geq$ TRF-RNN + LM + NSSP & 99\%\textsuperscript{*} & 99\%\textsuperscript{*}   \\
\end{tabular}
\end{center}
\end{table*}

To validate that making progress on the ROUGE metrics correlates to making real progress
as judged by humans, we conducted several side-by-side model comparisons
covering the range of ROUGE scores from
low to high-end
using Amazon Mechanical Turk workers. Workers were presented with the 5-sentence story/paragraph along with 
two model summaries and asked to rate each on two dimensions: fluency and information relevance. To minimize
inter-rater noise, scores by 3 distinct workers were collected for each example and averaged.
We aggregated results across 100 random examples from the test set.
Results showing the average preference of the workers on the two dimensions are presented in Table~\ref{tab:human}.

We observed that the fluency improved significantly from the MeanSum-single model through the RNN-RNN+critic+LM models, while the information aspect continued to improve through our best model, the TRF-RNN with NSSP and language model pre-training.
The human performance was still far better on both dimensions when compared side-by-side.
Additional model samples can be viewed in Appendix Figures~\ref{fig:parallel} and ~\ref{fig:good_samples}.

In addition to decoding a sentence from a paragraph representation, 
we found it informative to look at reconstructed paragraphs from the auto-encoder which are also included in  Figure~\ref{fig:good_samples}.
The paragraph reconstructions show some coherence, although with some 
disfluencies and factual inaccuracies that
are common with neural generative models.
Since the summaries are decoded from the same latent vector as the reconstructions, improving them could lead to more accurate summaries.

\section{Conclusions}
We introduce, ROCSumm, a new benchmark task for zero-shot, unsupervised abstractive summarization (UAS) that is useful for iterating and measuring progress on this challenging problem.
As one of the very first works approaching single-document UAS, we propose a novel neural model---\modelname---based on a denoising auto-encoder along with several self-supervised pre-training techniques for enhancing the model.
While performance is still far behind humans, \modelname\ outperforms extractive baselines, and is a major step toward UAS.

\section{Code and data release}
The code to reproduce our experimental setup is provided at \codeurl.
We include the code to process the ROCStories dataset into ROCSumm, our train/test splits,
the 3000 human summaries used in validation and test evaluation, Amazon Mechanical Turk templates used for data collection and evaluation. We also include
model training and evaluation code, and the model weights for our best model.

\bibliography{main}
\bibliographystyle{acl_natbib}

\appendix

\section{Appendix}
\label{sec:appendix}
\subsection{Amazon Mechanical Turk settings}
\label{app:amt}

\begin{figure*}[t]
	\centering
		\includegraphics[width=0.8\textwidth]{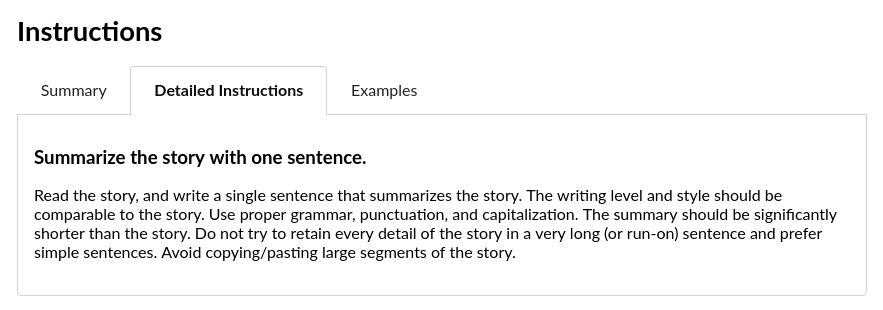}
		\includegraphics[width=0.5\textwidth]{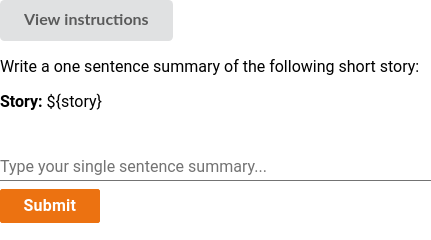}
	\caption{AMT task instructions.}
	\label{fig:collect_instructions}
\end{figure*}

Figure~\ref{fig:collect_instructions} shows the detailed instructions
for the ROCStories summarization task and the task user interface. We rewarded $\$0.15$ per summary
and restricted worker requirements to the following:
\begin{itemize}
	\item HIT Approval Rate $\%>98$
	\item Location: United States
	\item Number of HITs Approved $>1000$
\end{itemize}

We used the same worker requirements for the human evaluation experiment discussed in
Section \ref{sec:human_eval}. 

The templates used are available at our code repository: \codeurl.

\subsection{Training the adversarially-regularized auto-encoder}
\label{app:critic}

\begin{figure*}[ht!]
    \begin{center}
        \includegraphics[width=0.7\textwidth]{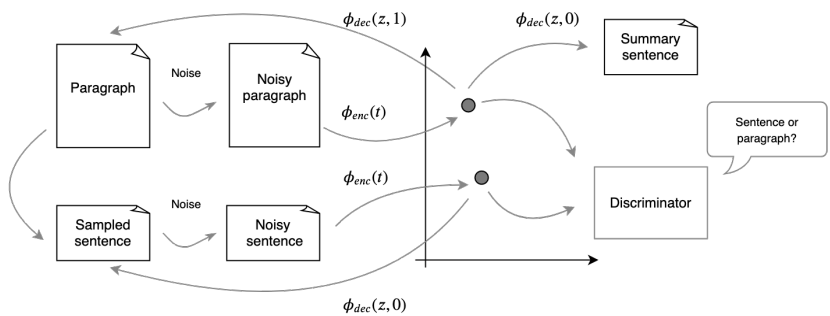}
        \caption{The auto-encoder model augmented with critic.}
        \label{fig:critic}
    \end{center}
\end{figure*}

The model architecture augmented with a critic is depicted in Figure \ref{fig:critic}.
The critic was implemented as a multi-layer perceptron~(MLP) with one hidden layer, parameterized by $\theta_D$, and trained with a standard classification loss:
\begin{equation}
  \ell_D(p, \theta_D) = -\mathbb{E}[p_D (y=y(t)| \enc(t), \theta_D)
  \label{eq:Dloss}
\end{equation}
where $t$ is  $p$ or a randomly sampled sentence.
The auto-encoder loss was augmented with  $\ell_{a}$ encouraging the encoder to fool $D$ into classifying paragraph vectors as sentences:
\begin{align}
	\ell_a(p, \encparam) =  -\mathbb{E}[p_D(y=0 | \enc(p))] \\
	\ell_{aae}(p, \encparam, \decparam)
	= \ell_{rec}(p, \encparam, \decparam) 
	+ \lambda_{a} \ell_{a}(p, \encparam)
    \label{eq:critic}
\end{align}
As with Generative Adversarial Networks \citep{goodfellow2014generative},
this is a two-player mini-max game, and in practice we alternate training the autoencoder (with respect to encoder and decoder parameters, $\encparam$ and $\decparam$) and the discriminator (with respect to $\theta_D$).
However, unlike GANs we do not generate data from the training domain,
in this case textual sequences, and instead operate in the continuous latent
space, which allowed us to maintain differentiability.

\begin{algorithm}

\SetKwData{Step}{step}
\SetKwData{Pretrain}{$N_{pretrain}$}
\SetKwInOut{Input}{Input}
\SetKwInOut{Output}{Output}
\Input{A collection of paragraphs, $P$}
 \For{$p$ in $P$}{
  \Step $\leftarrow$ \Step $+ 1$\\
  \uIf{\Step $<$ \Pretrain}{
      Update $(\encparam, \decparam)$ by gradient descent on $\ell_{rec}(p, \encparam, \decparam)$ (Eq.\ref{eq:rec})
  }\uElseIf{\Step$\bmod 2 = 0$}{
   Update $\theta_D$ by gradient descent on $\ell_{D}(p, \theta_D)$ (Eq.\ref{eq:Dloss})
   }
   \Else{
      Update $(\encparam, \decparam)$ by gradient descent $\ell_{ae}(p, \encparam, \decparam)$ (Eq.\ref{eq:rec})
   }
  }
  \Output{Auto-encoder parameters: $(\theta_E, \theta_G)$}
\caption{Our model training loop with critic}
\label{algo}
\end{algorithm}


\begin{figure*}[ht]
\small
\begin{mdframed}
\textbf{Story (paragraph) to summarize}: \\
Ben decided to go for a walk. \\
But he stepped outside and shivered. \\
The weather had turned cold overnight! \\
Ben went back in and put on another sweater. \\
Now he was ready for his walk! \\

\textbf{Human (from 3 different workers)}\\ 
- Ben put on more layers after noticing the cold and went on a walk. \\
- Ben went outside and realized it was cold so he put a sweater on. \\
- Ben put on more layers after noticing the cold and went on a walk. \\

\textbf{MeanSum-single + Noise}\\ Ben decided back in to also take him. \\\

\textbf{RNN-RNN + critic}\\ Ben decided to go for an clean glass. \\

\textbf{RNN-RNN + critic + LM}\\ Ben decided to go outside and he was in a dark! \\

\textbf{TRF-RNN + NSSP + LM (best model)}\\ Ben decided to go outside for awhile and he was ready for a ride back on and shivering!

\end{mdframed}

\normalsize

\caption{For the same story, we show summaries produced by different models used in human evaluation (Table~\ref{tab:human}).}
\label{fig:parallel}

\end{figure*}
\begin{figure*}[t]
  \small

  \begin{mdframed}
    \underline{\textbf{Sample \#1}}\\
    \textbf{Story:} Bill was always bullied. His bully was a star athlete. One day Bill pushed him back. The bully fell wrong and broke his ankle. It ruined his athletic career.\\
    \textbf{Summary:} Bill was always bullied by a bully and he broke his ankle but it broke.\\
    \textbf{Reconstruction:} Bill was always bullied. One day a bully was bullied. Bill beat his opponent. The ball broke down. Bill was crushed his ankle laced.\\

    \underline{\textbf{Sample \#2}}\\
    \textbf{Story:} Antonio was happy. His new pizza place was doing great business. His pizza place had an edge on all the others. He only used the freshest and best ingredients. His pizza place would be there for a long time.\\
    \textbf{Summary:} Antonio was happy to get his best pizza place and had a great time at the place.\\
    \textbf{Reconstruction:} Antonio was happy to do the best pizza place. is favorite place had a great time. His best friend had an extra place. His place was the best pizza place. His place had a good time with the pizza.\\

    \underline{\textbf{Sample \#3}}\\
    \textbf{Story:} Charles was sure he wouldn't qualify for an auto loan. He was 36 years old and didn't have a car. On a lunch break, he met a Nissan car salesman. The salesman invited Charles to apply for a car loan. Charles was approved and the salesman sold him a Nissan.\\
    \textbf{Summary:} Charles was a salesman and told him he needed a loan from a dealership to buy a car and was approved.\\
    \textbf{Reconstruction:} Charles was a salesman and wanted to quit. Charles applied for a job and wanted a car. Charles applied for a loan and asked him to pay a loan. Charles was offered a loan and asked him to quit. Charles was a salesman and told him he needed a loan.\\

    \underline{\textbf{Sample \#4}}\\
    \textbf{Story:} Sue wasn't feeling well. She had stayed in bed all day. Sue hadn't eaten all day. She barely drank anything. Sue got dehydrated.\\
    \textbf{Summary:} Sue wasn't feeling well in bed because she had barely drank after all.\\
    \textbf{Reconstruction:} Sue wasn't feeling well. She had never drank before. She had all night. Sue got all cold. Sue felt very refreshed.\\

    \underline{\textbf{Sample \#5}}\\
    \textbf{Story:} My boss asked me to find the mean value of a data set. I told him the mean was not appropriate for the data set. He asked me to explain why it was not appropriate. I told him it was because the data was not symmetrical. He thanked me for my brilliant insight into the problem.\\
    \textbf{Summary:} My boss asked me to make the mistake of the data entry for me, so I was pleased to find it.\\
    \textbf{Reconstruction:} My boss asked me to make the mistake of the data entry. He told me what the expectation was making him a special name. He told me it was the wrong size. He told me not to make the decision to be careful. I told me it was the perfect decision for me.\\

    \underline{\textbf{Sample \#6}}\\
    \textbf{Story:} Bob just got a new puppy. Bob's cat did not like the new puppy. They fought with each other constantly. Eventually Bob's pets became more familiar with each other. Bob's new puppy and cat eventually became best friends.\\
    \textbf{Summary:} Bob just got a new puppy to the other cat and was very happy with their new puppy.\\
    \textbf{Reconstruction:} Bob just got a new puppy. They were very cute and a little girl. They got the best cat they'd met with other kids. They eventually got a new puppy. The other person became very lonely now.\\

    \underline{\textbf{Sample \#7}}\\
    \textbf{Story:} Nathan wanted to be a computer programmer. He would code everyday after school. Eventually he got a degree in computer science. He was hired by Microsoft to help program their new operating system. Nathan was a big contribution to Microsoft's development team.\\
    \textbf{Summary:} Nathan wanted to be a great job at all his company's computer and was able to improve his new role.\\
    \textbf{Reconstruction:} Nathan wanted to be a programmer. He worked hard to get a new company. He was hired to improve his computer. He was hired to sponsor a month and was able to own a company. He was able to improve his own company.\\

    \underline{\textbf{Sample \#8}}\\
    \textbf{Story:} Fred has a fever. He was not able to go to work. He had to call out. His work boss was angry. He got fired.\\
    \textbf{Summary:} Fred has a fever and was not able to go to work out of work and got fired.\\
    \textbf{Reconstruction:} Fred has a fever. He was not able to go to work. He got out of work. He was mad. He was able to get out.
  \end{mdframed}

  \normalsize
  \caption{Some summaries generated by our best \modelname\ model along with the reconstructed stories.}
  \label{fig:good_samples}
\end{figure*}

\end{document}